\documentclass[letter]{article}
\usepackage{hyperref}
\hypersetup{
    colorlinks=true,
    linkcolor=blue,
    filecolor=magenta,      
    urlcolor=cyan,
}
\usepackage{tabularx}
\usepackage{INTERSPEECH2021}
\usepackage{enumitem}
\title{RyanSpeech: A Corpus for Conversational Text-to-Speech Synthesis}
\name{Rohola Zandie$^{1,2}$, Mohammad H. Mahoor$^{1,2}$, Julia Madsen$^2$, and Eshrat S. Emamian $^2$}
\address{
  $^1$Department of Electrical and Computer Engineering, University of Denver\\
  $^2$DreamFace Technologies, LLC}
\email{rohola.zandie@du.edu, mmahoor@du.edu, \{jmadsen, eemamian\}@dreamfacetech.com}

\begin{document}

\maketitle
\begin{abstract}
This paper introduces \textit{RyanSpeech}, a new speech corpus for research on automated text-to-speech (TTS) systems. Publicly available TTS corpora are often noisy, recorded with multiple speakers, or lack quality male speech data. In order to meet the need for a high quality, publicly available male speech corpus within the field of speech recognition, we have designed and created \textit{RyanSpeech} which contains textual materials from real-world conversational settings. These materials contain over 10 hours of a professional male voice actor's speech recorded at 44.1 kHz. This corpus's design and pipeline make \textit{RyanSpeech} ideal for developing TTS systems in real-world applications. To provide a baseline for future research, protocols, and benchmarks, we trained 4 state-of-the-art speech models and a vocoder on \textit{RyanSpeech}. The results show 3.36 in mean opinion scores (MOS) in our best model. We have made both the corpus and trained models for public use.

\end{abstract}

\section{Introduction}
\label{introduction}
\vskip -0.5em
The advent of end-to-end deep neural networks (DNN) in the fields of automatic speech recognition (ASR) and text-to-speech (TTS) has shifted the research paradigm in these fields. Traditional methods are complex and time-consuming, requiring pre-defined linguistic features which are typically language specific. However, new TTS architectures are composed of two parts: first, they generate a Mel-spectrogram autoregressively or non-autoregressively from the input text's phonemes, and then output audio from a separately trained vocoder. The quality of the final outputs depends heavily on the quality of speech corpora. Even though researchers \cite{ljspeech17, zen2019libritts, panayotov2015librispeech}  in recent years have paid more attention to create larger speech and language corpora, it is evident that more research is necessary to fill the gaps of speech corpora.

This paper addresses shortcomings in recent research and scholarship within the fields of corpus development by introducing \textit{RyanSpeech}, the first publicly available male voice TTS corpus in the conversational setting. Using state-of-the-art deep neural networks, we have found that it is easier to train a model capable of generating natural speech synthesizers than traditional methods \cite{zen2019libritts}. However, this is only possible when we have an available corpus at hand. Most of the publicly available corpora are in the domains of reading or audiobooks, and most of these corpora are not single-speakers, which is unsuitable for training the TTS (especially the vocoder models). To have a high-fidelity TTS, the corpus also needs to be recorded at a high sampling rate without any background noise. LJSpeech is the only female single-speaker corpus with low noise and a nonrestrictive license \cite{ljspeech17}. \textit{RyanSpeech} includes features that make it ideal for a TTS system in real-world applications. These features are:

\begin{itemize}
    \item \textit{RyanSpeech} is the only corpus in the domain of conversation. Other speech corpora in the domain of conversation are multi-speaker and recorded in high-noise environments that are not appropriate for training TTS systems. In \textit{RyanSpeech}, we extracted the most commonly used conversations with prosodic variations specific to conversations that cannot be replaced by audiobooks or reading settings. 
    
    \item All the audio files are recorded by a single professional male speaker with studio quality at a sampling rate of 44100 Hz. We double-checked all the recordings and rerecorded sample files that were too slow, too fast, or contained unwanted speech variations. This makes \textit{RyanSpeech} the first male speaker corpus for building high-quality TTS systems.
    
    \item We chose the sentences to reflect numerous and diverse real-life conversational situations, including dialogues on movies, sports, music, television, restaurants, and nature as well as common questions and general discussions. 
    
    \item We open-sourced the corpus including all the original FLAC (Free Lossless Audio Codec) files with transcripts under the \textit{CC BY-NC-ND} license to help rapid development in TTS research.
\end{itemize}

This paper is organized as follows. Section \ref{sec:related_work} summarizes speech corpora, especially those most relevant to our own. Section \ref{sec:corpus_creation_pipeline} details the corpus creation pipeline from collecting raw text to the final audio files. Section \ref{sec:statistics_of_the_corpus} provides the corpus's overall statistics, including both audio and textual transcripts. We present the details of training different models based on \textit{RyanSpeech} as well as the human evaluation and the results in Section \ref{sec:experiments}. Finally, we conclude the paper in Section \ref{sec:conclusions}. 

\section{Related Work}
\label{sec:related_work}
\vskip -0.5em
Speech corpora can be categorized into multi- and single-speakers. Multi-speaker corpora are designed to capture the diversity in spoken language with different voices, genders, ages, and accents. These corpora are well-suited to automatic speech recognition systems that require variations in input data \cite{acero2012acoustical}. However, creating a high-quality TTS system necessitates single-speaker corpora \cite{zen2019libritts}, which is especially important for training the vocoder.

Table \ref{tab:corpora} demonstrates different corpora with public licenses widely used for research on speech recognition and synthesis. The CMU ARCTIC corpus is a dataset that has been used for years as a baseline for speech recognition tasks \cite{kominek2003cmu}. The VCTK corpus contains 110 English speakers with various accents \cite{veaux2016superseded}. The main sources for the passages are newspapers selected using a greedy algorithm to increase phonetic diversity. Common Voice is a project by Mozilla which collects the biggest speech corpus by crowdsourcing from their community \cite{ardila2019common}. This corpus has the greatest diversity, and is also very noisy. VoxForge is most similar to Common Voice because it is also a community-based project \cite{VoxForge}. Unlike Common Voice, VoxForge does not have a verification process in the data collection process. LibriSpeech \cite{panayotov2015librispeech} and LibriTTS \cite{zen2019libritts} derive their text source from the LibriVox project which is based on audiobooks \cite{kearns2014librivox}. LibriTTS is the successor of LibriSpeech and has more robust design choices including high-quality sampling rate, the removal of noisy subsets, and split speech at sentence breaks. In the domain of conversational speech corpora, we have CHiME-5 \cite{barker2018fifth} which consists of 20 parties each recorded in different houses. Also, the Linguistic Data Consortium (LDC) has developed CALLHOME American English Speech, which consists of 120 unscripted 30-minute telephone conversations between native English speakers \cite{canavan1997callhome}. Both of these corpora are noisy and contain multiple speakers, which renders them not particularly suitable for TTS training.

As we mentioned above, single-speaker corpora are the best candidates for high-quality TTS systems. The BC2013 corpus comes from the blizzard challenge and contains a large amount of speech by a single speaker with good quality \cite{hu2017ustc}. While this corpus is based on Audiobooks, it is not ideal for many applications that require user interaction. 
M-AILABS was recorded with a male and female voice, but it is too noisy and unsuitable for training the speech synthesizer \cite{M-AILABS}. The only acceptable quality speech TTS corpus is LJSpeech, which is all recorded by a female voice. \textit{RyanSpeech} is the first high-quality male TTS corpus with a non-restrictive license in the conversational domain that can be used for training both the vocoder and speech model. 

\begin{table*}[t]
\caption{List of the publicly available multi and single speaker speech dataset}
\label{tab:corpora}
\begin{tabular}{cccccc}
\hline
Corpus              & Domain                           & Licence        & Duration (hours) & Sampling rate (kHz) & Number of Speakers   \\ \hline
CMU ARCTIC \cite{kominek2003cmu}           & Reading                          & BSD            & 7                & 16                  & 7                    \\
VCTK \cite{veaux2016superseded}                & Reading                          & ODC-By v1.0    & 44               & 48                  & 109                  \\
Common Voice \cite{ardila2019common}        & Reading                          & CC-0 1.0       & 1,118            & 48                  & 51,072               \\
VoxForge    \cite{VoxForge}        & Reading                          & GPL            & 120              & 16                  & 2966                 \\
LibriSpeech \cite{panayotov2015librispeech}        & Audiobook                        & CC-BY 4.0      & 982              & 16                  & 2,484                \\
LibriTTS \cite{zen2019libritts}            & Audiobook                        & CC-BY 4.0      & 586              & 24                  & 2,456                \\
CHiME-5 \cite{barker2018fifth}             & \multicolumn{1}{l}{Conversation} & Commercial     & 50               & 16                  & 48                   \\
CALLHOME \cite{canavan1997callhome}            & Conversation                     & LDC            & 60               & 8                   & 120 \\
BC2013 \cite{hu2017ustc}              & Audiobook                        & Non-commercial & 300              & 44.1                & 1                    \\
M-AILABS \cite{M-AILABS}            & Audiobook                        & BSD            & 75               & 16                  & 2                    \\
LJSpeech \cite{ljspeech17}            & Audiobook                        & CC-0 1.0       & 25               & 22.05               & 1                    \\
\textbf{RyanSpeech} & Conversation                     & CC BY-NC-ND               & 10               & 44.1                & 1                    \\ \hline
\end{tabular}
\end{table*}

\section{Corpus creation pipeline}
\label{sec:corpus_creation_pipeline}
\vskip -0.5em
This section describes the data processing which we developed to produce the \textit{RyanSpeech} corpus. 
\subsection{Data collection}
\vskip -0.5em
Since \textit{RyanSpeech} is a single-speaker speech corpus designed and developed for conversational systems, we used three different text resources that are most relevant to this task.

\begin{enumerate}[leftmargin=*]
\item Ryan Chatbot dataset: This is a specifically designed dataset developed for Ryan Robot \cite{ryanstudy2021, Francesca2019} in different conversational settings. It contains more than 56,000 sentences covering a wide variety of topics, including television, sports, movies, music, science, food, museums, and history. We randomly selected 5778 sample sentences from this dataset, and what follows is a short list of example sentences that appear in our “food” dialogues:
\begin{itemize}[leftmargin=*]
\item “Do you want to see another recipe that you could easily prepare at home?”
\item “I'm coming over to your house, and I'm coming hungry!”
\item “Maybe you just need to have a nice, home-cooked meal.”
           % \item “Would you say you prefer Chinese, Thai, or Indian food?”
\end{itemize} 
\item Taskmaster-2: This dataset is also designed for use in goal-oriented dialogue systems \cite{byrne2019taskmaster}. It includes both user and assistant roles in the conversations. Taskmaster-2 consists of 17289 dialogues in the seven domains of restaurants, food ordering, movies, hotels, flights, music, and sports. For \textit{RyanSpeech}, we randomly selected 3000 samples in all categories of Taskmaster-2.
\item LibriTTS: To balance the dataset, we also included 2501 random samples from the LibriTTS text corpus. LibriTTS is a multi-speaker speech dataset with 116500 sentences in their train-clean-360 subset. We randomly selected 2501 sentences from this subset.
\end{enumerate}

\subsection{Text pre-processing}
\vskip -0.5em
After collecting texts from different sources, we underwent sentence segmentation and normalization.
\begin{enumerate}[leftmargin=*]
    \item Sentence segmentation: We used Spacy \cite{spacy} for sentence segmentation on text from the Ryan Chatbot dataset and Taskmaster-2. It uses a trainable pipeline component for sentence segmentation.
    
        \item Text normalization: we detected non-standard words and normalized them to read text manually. In this step, we address numbers (cardinal numbers, signed integers, real numbers, ordinal numbers, roman numerals, fractions, and sequence of digits like phone numbers), Currency, Time, Date, Abbreviations and Street addresses.
    % \begin{itemize}
    %     \item Numbers: cardinal numbers, signed integers, real numbers, ordinal numbers, roman numerals, fractions, and sequence of digits like phone numbers.
    %     \item Currency: all different currencies have been addressed.
    %     \item Time: time specified both in 12-hour and 24-hour.
    %     \item Date: Standard US format (M/D/Y)
    %     \item Abbreviations
    %     \item Street addresses
    % \end{itemize}
    
\end{enumerate}

\subsection{Trim silence}
\vskip -0.5em
After recording the audio files, we needed to trim the beginning and end silences in each file. For this purpose, we wrote a simple script to trim the audio based on thresholding the sound amplitude. To make it even more accurate, we manually double-checked all the trimmings with Audacity audio software to ensure that we correctly removed the silences.

\subsection{Post processing}
\vskip -0.5em
Even though all the audios had been recorded in a studio with high-quality, we normalized the sound amplitude of recordings to ensure that they all had the same volume level.

\section{Statistics of the corpus}
\label{sec:statistics_of_the_corpus}
\vskip -0.5em
\textit{RyanSpeech} contains 9.84 hours of high-quality audio recorded in a studio by a professional voice actor. The speaker has the US English dialect. The sampling rate of all audio files is 44,100 Hz. We used the audio coding format of FLAC, which is lossless compression.

\begin{figure}[]
                \centering
            \includegraphics[width=6cm]{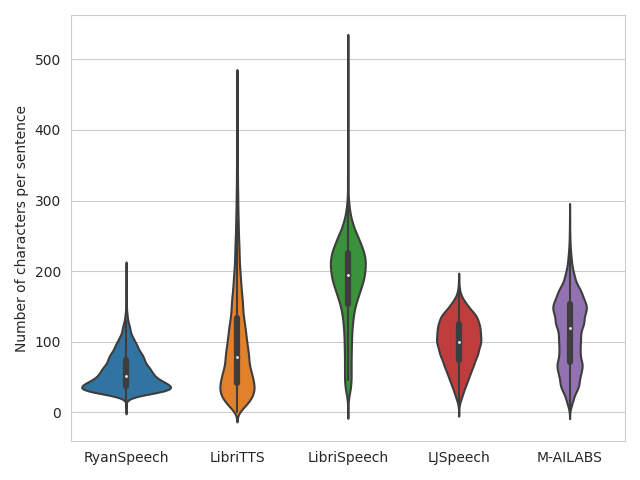}
                \caption{The violin plots of the number of characters per sentence for \textit{RyanSpeech}, LibriTTS, LibriSpeech, LJSpeech and, M-AILABS. The thick line shows the interquartile range (from 25\% to 75\%), and the white dot is the median value. The width of the violin plot in any point indicates the frequency.}
                \label{fig:violin_plot}
\vskip -1.5em
\end{figure}

Figure \ref{fig:violin_plot} shows violin plots of the number of characters per sentence in LibriTTS, LJSpeech, LibriSpeech, M-AILABS, and \textit{RyanSpeech}. It can be seen from the figure that \textit{RyanSpeech} has relatively shorter sentences compared to other corpora. The mean sentence length is 58.07 characters with a standard deviation of 26.09, closest to LibriTTS. The distribution of sentence length from \textit{RyanSpeech} to M-AILABS is also significantly different. Our corpus sentence lengths follow a peaked distribution, whereas in M-AILABS, it has a larger standard deviation (std=50.80).

\begin{figure}[t]
                \centering
              \includegraphics[width=6cm]{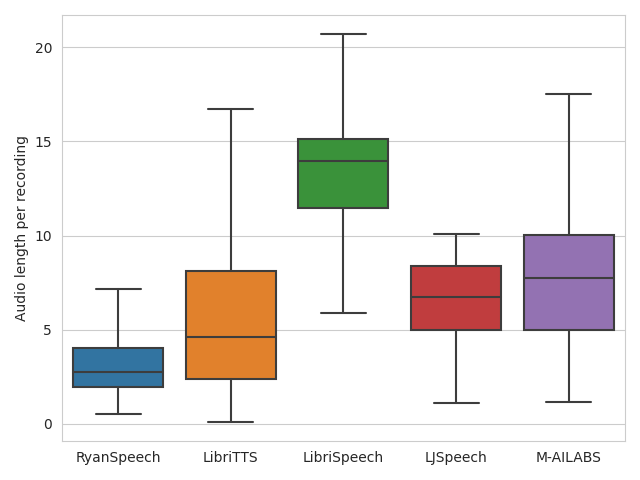}
                \caption{The box plots of audio length for \textit{RyanSpeech}, LibriTTS, LibriSpeech, LJSpeech and, M-AILABS. The box shows the interquartile range (from 25\% to 75\%); the middle line shows the median value. The whole range is from the minimum to maximum value. \textit{RyanSpeech} has the shortest audio length which is a characteristic of conversational dialogue.}
                \label{fig:box_plot}

\end{figure}

In Figure \ref{fig:box_plot}, the Box Plot of audio duration for different corpora are demonstrated. Because our corpus is in the domain of conversation, the audio files are shorter. The main reason for this observation is that the properties of conversational recordings are usually shorter and faster than audiobooks and other reading corpora. The diversity of audio length in terms of standard deviation in \textit{RyanSpeech} is less than other corpora, making it suitable for a stable training corpus for assistant and chat dialogue systems.

\section{Experiments}
\label{sec:experiments}
% This section presents experimental results on training the vocoder and TTS models based on the \textit{RyanSpeech} corpus to provide the baselines. 
\vskip -0.5em
The training pipeline consists of two steps: first, the character level input text is fed into a TTS system that outputs Mel spectrogram frames, then we use a trained vocoder to convert the Mel spectrograms to waveforms.

\subsection{Training neural vocoder}
\begin{figure}[t]
\centering
\includegraphics[width=8cm]{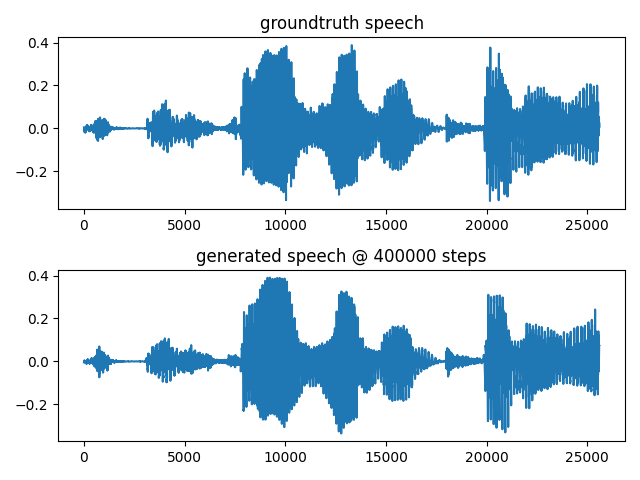}
\caption{The plot for the audio generated by the ParallelWaveGAN vocoder after training at 400k steps versus the ground truth.}
\label{fig:parallelwavegan_sample}
\end{figure}

\vskip -0.5em
The speech synthesizer gives us the time-domain waveform samples from the Mel spectrograms feature representations produced by the TTS system. In the first step, we trained ParallelWaveGAN \cite{yamamoto2020parallel}, which is a fast waveform generation method that uses a non-autoregressive WaveNet model. It is based on Generative Adversarial Network (GAN) and does not need a two-stage teacher-student framework for training. Basically, the model is trained by optimizing multi-resolution Short-time Fourier transform (STFT) on the spectrogram and an adversarial loss function at the same time.
For training the feature extractor, the sampling rate is set to 22,050 Hz with the FFT window size of 1024. For the generator, we set the number of residual block layers to 30, the number of dilation cycles to 3, and the number of residual channels to 64. The number of layers for the discriminator network is set to 10. The loss balancing coefficient is also set to 4. The model was trained for 400k steps. Figure \ref{fig:parallelwavegan_sample} shows one example of the generated sample after 400k steps versus the ground truth. It illustrates that the generative network is capable of creating a very similar waveform to the ground truth. The training was done on a NVIDIA TITAN-Xp GPU with 12 GB of memory and a batch size of 6. The training takes 95.74 hours. 
% There are different kinds of losses that need to be optimized. After training, the adversarial loss was 0.2547, the discriminator loss was 0.4952, and the generator loss was 1.9417. 

\begin{table}[t]
\centering
\caption{The comparison between standard deviation ($\sigma$), skewness($\gamma$) and kurtosis ($\kappa$) of pitch in ground-truth and synthesized audio}
\label{tab:pitch_results}
\begin{tabular}{cccc}
\hline
Model       & $\sigma$ & $\gamma$ & $\kappa$ \\ \hline
Ground Truth & 65.64             & -0.630   & 0.256   \\ \hline
Tacotron     & 63.97             & \textbf{-0.677}   & \textbf{0.302}   \\
FastSpeech   & 68.61             & -0.532   & 0.165  \\
FastSpeech2  & 67.71             & -0.565   & -0.023   \\
Conformer    & \textbf{65.77}             & -0.703   & 0.119   \\ \hline
\end{tabular}
\vskip -1.5em
\end{table}

\subsection{Training text to speech model}
\vskip -0.5em
We used different architecture for training the TTS system. Tacotron is a RNN-based model that uses CBHG (1-D convolution bank + highway network + bidirectional GRU) module \cite{wang2017tacotron}. We trained Tacotron for 100k steps. Transformer-based models have been adopted in TTS systems due to their success in modeling long-range dependencies. We trained FastSpeech \cite{ren2019fastspeech} and FastSpeech2 \cite{ren2020fastspeech} which are based on transformer. FastSpeech is based on attention mechanism and 1D convolution. These models have a module for length regulation that addresses the problem of mismatch between the phoneme and spectrogram sequence. The main difference between FastSpeech and FastSpeech2 is that the duration predictor for length regulation is trained using a teacher model in FastSpeech, while in FastSpeech2 it is trained end to end, which is faster and more accurate. Finally, we used Conformer \cite{gulati2020conformer} which brings the ideas of convolutional networks to the transformer-based model. The main difference between conformer and transformer models is that conformer has two feedforward layers which sandwiches not only a multi-head self-attention module but also a convolution module. The convolution module itself has a pointwise convolution projecting into a glue activation layer, followed by 1-D depthwise convolution. Finally, there is swish activation, another pointwise convolution, and the  batch norm. We trained FastSpeech, FastSpeech2, and Conformer for 500k steps.

We used ESPNet-TTS \cite{hayashi2020espnet}, which is an open-source end-to-end TTS toolkit that has implemented various state-of-the-art models. It provides recipes that are inspired by the Kaldi ASR toolkit. The training for all models was done on one NVIDIA TITAN Xp GPU with a batch size of 20. Table \ref{tab:training} depicts the comparison between different models in terms of the number of epochs, training time, and inference speed. Note that training time is only for the acoustic model, not including vocoder training. Except for Tacotron, the other models have the same inference speed. Conformer takes a longer time to train, but the quality is superior to other models. All of our pretrained models are free to download along with the original dataset.\footnote{code:\url{https://github.com/roholazandie/ryan-tts}}\footnote{corpus:\url{http://mohammadmahoor.com/ryanspeech/}} We refer the readers to \cite{hayashi2020espnet} for details of hyper-parameters and training configurations for each model. 

\begin{table}[]
\centering
\caption{The comparison of training epochs, training time and inference latency synthesis in different models. RTF (the real-time factor) denotes the time (in seconds) required for the system to synthesize one second waveform. The inference is done on NVIDIA GeForce GTX 1080 with 8GB and training on NVIDIA TITAN  Xp GPU with 12GB}
\label{tab:training}
\begin{tabularx}{\columnwidth}{XXXX}
\hline
            & Num epochs & Training time (h) & Inference speed (RTF) \\ \hline
Tacotron    & 200        & 49.58             & 0.12331               \\
FastSpeech  & 1000       & 26.19             & 0.04349               \\
FastSpeech2 & 1000       & 26.20             & 0.04355               \\
Conformer   & 1000       & 66.71             & 0.04516               \\ \hline
\end{tabularx}
\end{table}

\begin{table}[]
\centering
\caption{Comparison of MOS (mean opinion score) among our trained models with 95\% confidence intervals.}
\label{tab:mos_results}
\begin{tabular}{p{3cm}p{3cm}}
\hline
model       & MOS  \\ \hline
Tacotron    & 3.00 $\pm$ 0.18  \\
FastSpeech  & 3.27 $\pm$ 0.18 \\
FastSpeech2 & 3.27 $\pm$ 0.18 \\
Conformer   & 3.36 $\pm$ 0.17 \\ \hline
\end{tabular}
\vskip -1.5em
\end{table}

\subsection{Results}
\label{sec:results}
\vskip -0.5em
We randomly selected 30 fixed text samples with various lengths from the test corpus for evaluation. Different models all used the same text so that testers only reviewed the audio quality without interference factors. We then used a total of 120 audio samples generated by all four models for human evaluation. For a fair comparison, all the systems used Parallel WaveGAN as a vocoder. Twelve human subjects (native American English speakers, 18 years of age or older) participated in this study. Each participant independently scored all the models' outputs that were randomly shuffled. The evaluation was based on Likert scale score (1: Very Poor, 2:Poor, 3: Fair, 4: Good, 5: Excellent).

Quantitative subjective evaluation, known as mean opinion score (MOS), has been used for results analysis \cite{chu2006objective}. MOS is simply the mean of the scores from all evaluators. Table \ref{tab:mos_results} demonstrates the difference between the evaluations on each model. The best subjective result on our models is the conformer model. FastSpeech and FastSpeech2 have the same MOS score in our evaluations. Due to the effectiveness of variance information such as frequency and energy, as well as increased accuracy for predictions of duration, FastSpeech models are superior to Tacotron. However, the conformer model achieved the best score (closer to good on Likert scale) because of the ability of the model to harness the power of CNNs in order to extract local features in the creation of the Mel-spectrogram. Our results are comparable to results from the same models trained on LJSpeech \cite{ren2019fastspeech}. 

To analyze the variance information, we calculated the first three moments of pitch distribution for the ground truth and synthesized speech \cite{andreeva2014differences, niebuhr2019measuring}. Table \ref{tab:pitch_results} shows the results. Among different models, Conformer is closer to the ground truth in terms of standard deviation, though Tacotron is closest with respect to skewness and kurtosis. This demonstrates that Conformer can produce pitch distribution close to natural voices, which results in better prosody.

\section{Conclusions and Outlook}
\label{sec:conclusions}
\vskip -0.5em
In this paper, we introduced the \textit{RyanSpeech} corpus, which we designed primarily for TTS systems. We collected the text dataset from a variety of sources with a particular focus on conversational settings, and designed the corpus to be high-quality in its studio settings. As we have demonstrated, this is the first large, high-quality male speech corpus that is open-source. \textit{RyanSpeech} can open up a number of possibilities for future research in the areas of ASR and TTS. With the trend in speech synthesis to find models that replicate human speech with small datasets, \textit{RyanSpeech} proves an excellent candidate, especially for TTS evaluation and development in research as well as applications that require natural sources of speech, such as movies and podcasts.

\section{Acknowledgements}
\vskip -0.5em
Research reported in this paper was supported by the National Institute on Aging of the National Institutes of Health under award number R44AG059483 to DreamFace Tech, LLC.
\bibliographystyle{IEEEtran}

\bibliography{main}

\end{document}